\documentclass[letterpaper,journal]{IEEEtran}
\usepackage{amsmath,amsfonts}
\usepackage{algorithmic}
\usepackage{algorithm}
\usepackage{array}
\usepackage[caption=false,font=normalsize,labelfont=sf,textfont=sf]{subfig}
\usepackage{textcomp}
\usepackage{stfloats}
\usepackage{url}
\usepackage{verbatim}
\usepackage{graphicx}
\usepackage{cite}
\usepackage{booktabs}
\usepackage{multirow}
\usepackage{tikz}
\usetikzlibrary{shapes.geometric, arrows, positioning, fit, calc}

\begin{document}

\title{MambaRefine-YOLO: A Dual-Modality Small Object Detector for UAV Imagery}

\author{Shuyu Cao\textsuperscript{*},~
        Minxin Chen\textsuperscript{*},~
        Yucheng Song,~
        Zhaozhong Chen,~
        and Xinyou Zhang
\thanks{*~S.~Cao and M.~Chen contributed equally to this work.
S.~Cao, M.~Chen, Y.~Song, and Z.~Chen are with the SWJTU-Leeds Joint School,
Southwest Jiaotong University, Chengdu, China(e-mail: shuyucao03@gmail.com, sztop2020@gmail.com, yucheng\_song2419@outlook.com); X.~Zhang is with the School of
Computing and AI, Southwest Jiaotong University, Chengdu, China (e-mail: xyzhang@swjtu.edu.cn).}}


\maketitle

\begin{abstract}
Small object detection in Unmanned Aerial Vehicle (UAV) imagery is a persistent challenge, hindered by low resolution and background clutter. While fusing RGB and infrared (IR) data offers a promising solution, existing methods often struggle with the trade-off between effective cross-modal interaction and computational efficiency. In this letter, we introduce MambaRefine-YOLO. Its core contributions are a Dual-Gated Complementary Mamba fusion module (DGC-MFM) that adaptively balances RGB and IR modalities through illumination-aware and difference-aware gating mechanisms, and a Hierarchical Feature Aggregation Neck (HFAN) that uses a ``refine-then-fuse'' strategy to enhance multi-scale features. Our comprehensive experiments validate this dual-pronged approach. On the dual-modality DroneVehicle dataset, the full model achieves a state-of-the-art mAP of 83.2\%, an improvement of 7.9\% over the baseline. On the single-modality VisDrone dataset, a variant using only the HFAN also shows significant gains, demonstrating its general applicability. Our work presents a superior balance between accuracy and speed, making it highly suitable for real-world UAV applications.
\end{abstract}


\section{Introduction}
\IEEEPARstart{O}{bject} detection using Unmanned Aerial Vehicles (UAVs) is a critical technology for applications ranging from precision agriculture to disaster response \cite{li2024sod}. However, detecting objects from aerial platforms presents unique and significant challenges. Due to high altitudes, objects of interest often appear as small, low-resolution targets, making them difficult to distinguish from complex backgrounds \cite{nikouei2025_small_object_detection, liu2020uav}. These issues, combined with drastic scale variations and challenging lighting, often lead to missed detections for standard detectors.


To overcome these limitations, fusing data from RGB and infrared (IR) sensors has become an effective strategy \cite{fusion2024survey}. While deep learning-based fusion methods have shown promise \cite{improving2024coarse}, designing an architecture that effectively merges these two data streams remains a key challenge. Many fusion methods built on Convolutional Neural Networks (CNNs) are limited by the local nature of convolutions \cite{efficient2023inductive}. While Vision Transformers (ViTs) excel at capturing global relationships \cite{dosovitskiy2021image}, their quadratic computational cost makes them ill-suited for real-time processing of high-resolution UAV imagery \cite{du2025evit}. This creates a pressing trade-off: how can we achieve global cross-modal interaction without sacrificing real-time performance?

To break this impasse, we turn to Mamba, a recently developed state-space model (SSM) that achieves the long-range modeling capabilities of Transformers but with linear computational complexity \cite{gu2023mamba}. Its selective scan mechanism is an ideal candidate for fusing complementary information from different modalities without the computational overhead \cite{zhu2024vision}.

In this letter, we propose \emph{MambaRefine-YOLO}. Our key innovations are twofold. First, we introduce a Dual-Gated Complementary Mamba (DGC-MFM)fusion module for efficient and effective feature fusion. Second, we design a Hierarchical Feature Aggregation Neck (HFAN) that employs a ``refine-then-fuse'' strategy to enhance multi-scale features. Our contributions are:
\begin{itemize}
    \item A Dual-Gated Complementary Mamba fusion module (DGC-MFM) that employs illumination-aware and difference-aware gating mechanisms to adaptively fuse RGB and IR features, while maintaining linear computational complexity.
    \item A Hierarchical Feature Aggregation Neck (HFAN) that refines features before fusion to improve multi-scale representation for small objects.
    \item A comprehensive experimental validation on both dual-modality and single-modality datasets, demonstrating the effectiveness of the fusion module and the general applicability of the HFAN.
\end{itemize}

\section{Proposed Method}

\begin{figure*}[!t]
\centering
\includegraphics[width=\textwidth]{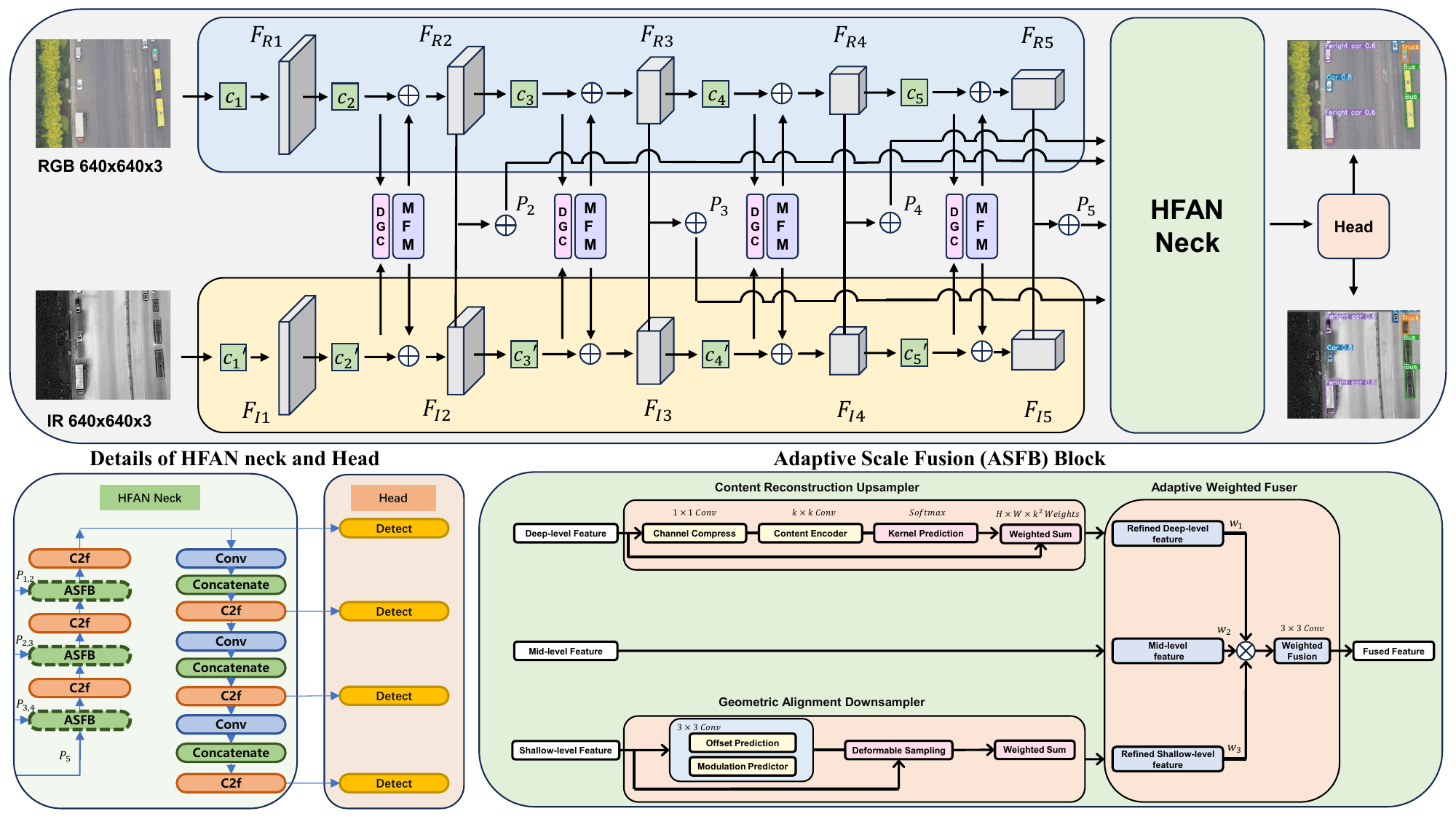}
\caption{The overall architecture of MambaRefine-YOLO. It consists of a dual-stream backbone where Dual-Gated Complementary Mamba Fusion Modules (DGC-MFM) are applied at four different scales ($C_2$ to $C_5$). The fused features are then processed by the Hierarchical Feature Aggregation Neck (HFAN), which contains several Adaptive Scale Fusion Blocks (ASFB). Finally, a multi-scale detection head produces the output.}
\label{fig:architecture}
\end{figure*}

Our proposed MambaRefine-YOLO is an end-to-end network designed to address two primary challenges in dual-modality UAV object detection: 1) achieving effective cross-modal fusion, and 2) enhancing multi-scale features to improve sensitivity to small objects. The overall architecture is depicted in Fig. \ref{fig:architecture}. It consists of two key components: a \emph{Dual-Stream Mamba-based Backbone} for feature extraction and fusion and a \emph{Hierarchical Feature Aggregation Neck (HFAN)} for feature refinement.

\subsection{Dual-Stream Mamba-based Backbone}

\begin{figure}[!t]
\centering
\includegraphics[width=\columnwidth]{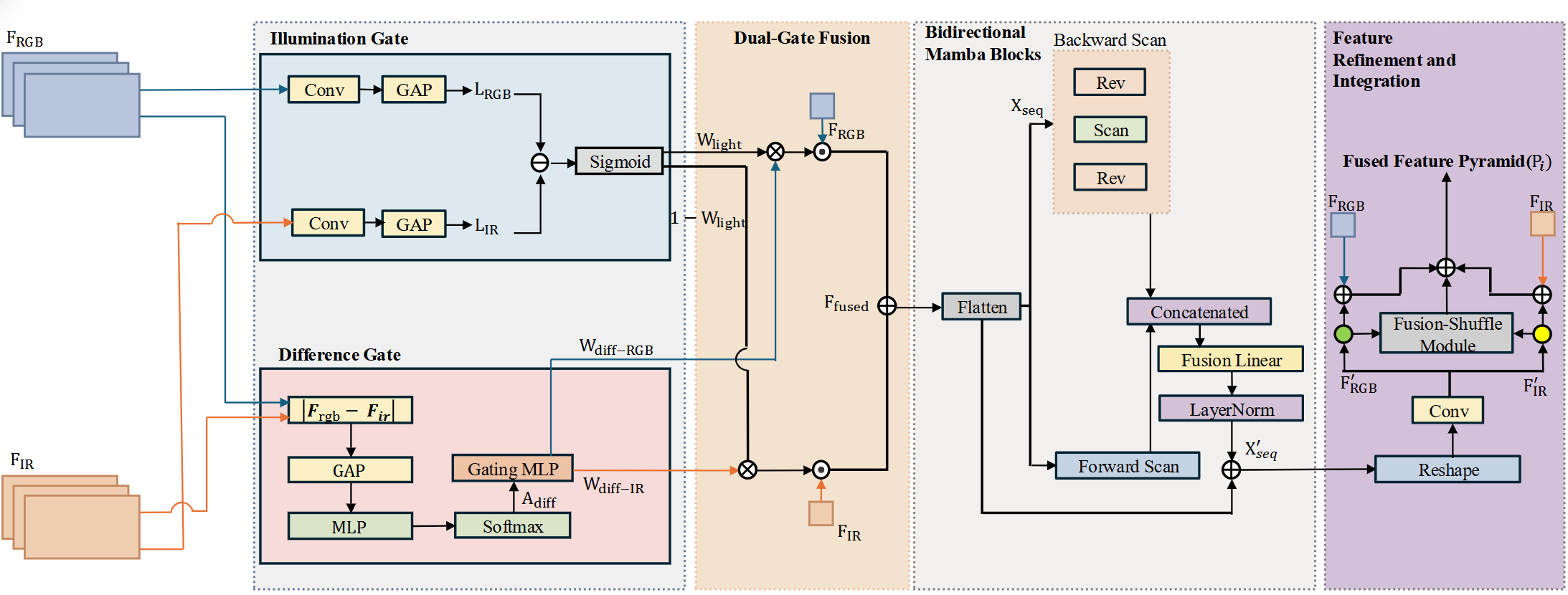}
\caption{The Dual-Gated Complementary Mamba Fusion Module (DGC-MFM) consists of four main stages: (1) Illumination Gate (IG) and Difference Gate (DG) generate adaptive weights, (2) Dual-gated fusion combines RGB and IR features, (3) Bidirectional Mamba processes the fused features to capture global context, and (4) Feature Refinement and Integration utilizes residual connections and a Fusion-Shuffle mechanism to generate the final feature pyramid.}
\label{fig:mfm_asfb}
\end{figure}

We employ a dual-stream backbone to integrate RGB and IR information. To achieve long-range cross-modal interaction while avoiding the high computational overhead or the restricted receptive fields of CNNs, we introduce the Dual-Gated Complementary Mamba Fusion Module (DGC-MFM) as illustrated in Fig.\ref{fig:mfm_asfb}. This module is applied at four feature scales ($C_2$ to $C_5$). Given feature maps $F_{\text{rgb}}, F_{\text{ir}} \in \mathbb{R}^{B \times C \times H \times W}$, the DGC-MFM operates as follows.

\noindent\textbf{1) Illumination Gate (IG).}
Inspired by illumination-aware weighting strategies~\cite{guan2019illumination}, we introduce an Illumination Gate (IG) to adjust modality weights according to lighting conditions. We first estimate the brightness of each modality using a convolution followed by Global Average Pooling (GAP), yielding scalar estimates $L_{\text{rgb}}$ and $L_{\text{ir}}$. An illumination weight is then computed by comparing these levels:
\begin{equation}
W_{\text{light}} = \sigma\big(\gamma(L_{\text{rgb}} - L_{\text{ir}})\big),
\label{eq:ig_Wlight}
\end{equation}
where $\sigma(\cdot)$ is the Sigmoid function and $\gamma$ is a learnable temperature parameter. The scalar $W_{\text{light}}$ is broadcast to the feature map dimensions, prioritizing RGB features in well-lit environments and IR features in low-light scenarios.

\noindent\textbf{2) Difference Gate (DG).}
To capture complementary information~\cite{jang2025mcor}, we propose a Difference Gate (DG). It models content-level discrepancies by computing the absolute difference between feature maps:
\begin{equation}
F_{\text{diff}} = \big|F_{\text{rgb}} - F_{\text{ir}}\big|.
\label{eq:dg_Fdiff}
\end{equation}
Then a channel attention vector $A_{\text{diff}}$ is derived via GAP to identify informative channels:
\begin{equation}
A_{\text{diff}} = \mathrm{Softmax}\big(W_2 \,\delta\big(W_1 \,\mathrm{GAP}(F_{\text{diff}})\big)\big),
\label{eq:dg_Adiff}
\end{equation}
where $W_1 \in \mathbb{R}^{d \times C}$ and $W_2 \in \mathbb{R}^{C \times d}$ are learnable parameters, and $\delta(\cdot)$ is a nonlinear activation. 
Using $A_{\text{diff}}$, the module generates modality-specific weights $W_{\text{diff-rgb}}$ and $W_{\text{diff-ir}}$. The final dual-gated fusion combines illumination and difference cues:
\begin{equation}
F_{\text{fused}} = \big(W_{\text{light}} \otimes W_{\text{diff-rgb}}\big) \odot F_{\text{rgb}}
                  + \big((1 - W_{\text{light}}) \otimes W_{\text{diff-ir}}\big) \odot F_{\text{ir}},
\label{eq:dgc_fused}
\end{equation}
where $\otimes$ denotes broadcast multiplication. This formulation integrates both illumination and content-level discrepancies into the fused representation, which serves as the input for the subsequent Bidirectional Mamba Blocks to model global dependencies.

\noindent\textbf{3) Bidirectional Mamba Backbone.}
To capture global context efficiently, the fused feature map $F_{\text{fused}}$ is flattened into a sequence $X_{\text{seq}} \in \mathbb{R}^{B \times N \times C}$ and processed by a stack of \emph{Bidirectional Mamba Blocks}. We utilize forward and backward state-space scans to model long-range dependencies with linear complexity $\mathcal{O}(N)$. The outputs from both directions are concatenated and fused via a linear projection:
\begin{equation}
X'_{\text{seq}} = \mathrm{LayerNorm}\Big(X_{\text{seq}} + \mathrm{Fusion}_{\text{bi}}\big(\mathrm{Cat}(Y_{\text{fwd}}, Y_{\text{bwd}})\big)\Big),
\label{eq:bi_mamba_update}
\end{equation}
where $Y_{\text{fwd}} = \mathrm{Mamba}_{\text{fwd}}(X_{\text{seq}})$, $Y_{\text{bwd}} = \mathrm{Rev}(\mathrm{Mamba}_{\text{bwd}}(\mathrm{Rev}(X_{\text{seq}})))$, and $\mathrm{Fusion}_{\text{bi}}$ is a learnable linear layer. This bidirectional approach eliminates the directional bias of causal modeling, ensuring that each feature token aggregates comprehensive global context essential for distinguishing targets from complex backgrounds.

\noindent\textbf{4) Feature Refinement and Integration.}
The output sequence $X'_{\text{seq}}$ is reshaped to the spatial domain and projected via a $1\times1$ convolution to generate two refined feature maps, $F'_{\text{rgb}}$ and $F'_{\text{ir}}$. To recover local details potentially smoothed during global modeling, we employ residual connections:
\begin{equation}
F^{\text{out}}_{m} = F_{m} + F'_{m}, \quad m \in \{\text{rgb},\text{ir}\}.
\label{eq:residual_fusion_compact}
\end{equation}
Subsequently, a \emph{Fusion-Shuffle Module} utilizing group convolutions and channel shuffling is applied to enhance cross-channel interaction. Finally, the features from both streams are summed to construct the multi-scale feature pyramid $\{P_2, P_3, P_4, P_5\}$ for the detection neck.

\begin{table*}[!t]
\centering
\caption{Comparison with state-of-the-art methods on the DroneVehicle dataset. All methods are evaluated using the mAP@.5 metric. 
         The best result is in \textbf{bold}, and the second best is \underline{underlined}.}
\label{tab:dronevehicle_sota_comprehensive}

\resizebox{\textwidth}{!}{%
    \renewcommand{\arraystretch}{1.1} 
    \tiny
    \begin{tabular}{l|l|cccccc}
    \toprule
    Method & Modality & Car & Truck & Freight-car & Bus & Van & mAP@.5 (\%) \\
    \cline{1-8}
    
    YOLO11 (Base) \cite{Github'24} & \multirow{2}{*}{RGB} & 96.4 & 74.4 & 54.2 & 95.0 & 56.3 & 75.3 \\
    Hu et al. (RS'23) \cite{hurs23} & & 96.2 & 75.8 & 57.3 & 94.5 & 56.7 & 76.1 \\
    
    \cline{1-8}
    
    YOLO11 (Base) \cite{Github'24} & \multirow{2}{*}{IR} & 98.3 & 77.5 & 65.8 & 95.0 & 59.9 & 79.3 \\
    Hu et al. (RS'23) \cite{hurs23} & & 98.0 & 79.5 & 67.2 & 94.8 & 58.6 & 79.6 \\
    
    \cline{1-8}
    
    CFT \cite{fang2021_cft} & \multirow{7}{*}{RGB+IR} &  97.0 & 79.5 & 64.0 & 78.0 & 66.0 & 76.9 \\
    M2FP (J-STARS'24) \cite{Ouyang2024M2FP} & & 95.7 & 76.2 & 64.7 & 92.1 & 64.7 & 78.7 \\
    Fusion-Mamba (TMM'24)\cite{Dong2024FusionMambaCross} & & 96.7 & 80.2 & 59.5 & 95.3 & 64.0 & 79.2 \\
    OAFA (CVPR'24) \cite{Chen_2024_CVPR} & & 90.3 & 76.8 & \textbf{73.3} & 90.3 & 66.0 & 79.4 \\
    UAVD-Mamba \cite{Li2025UAVD-Mamba} & & \textbf{98.6} & \textbf{83.9} & 69.8 & \underline{96.9} & \underline{66.1} & \underline{83.0} \\
    
    \cline{1-1} \cline{3-8}
    
    \textbf{MambaRefine-YOLO (Ours)} & & \underline{98.4} & \underline{79.5} & \underline{72.3} & \textbf{97.3} & \textbf{68.5} & \textbf{83.2} \\
    \bottomrule
    \end{tabular}%
}
\end{table*}

\subsection{Hierarchical Feature Aggregation Neck (HFAN)}
\label{sec:hfan}

Conventional feature pyramid networks often struggle with UAV imagery, as the direct fusion of multi-scale features leads to spatial misalignment and semantic conflicts, severely hindering small object detection. To address this, we propose a \emph{“Refine-then-Fuse” strategy}. This approach prioritizes targeted feature refinement to ensure informational quality and spatial consistency prior to fusion. Our Hierarchical Feature Aggregation Neck (HFAN) is designed to implement this principle.

The core of HFAN is the \emph{Adaptive Scale Fusion Block (ASFB)} (Fig.~\ref{fig:architecture}), which implements the proposed strategy via three key modules.

\noindent\textbf{1) Content Reconstruction Upsampler.} Deep features typically suffer from the loss of high-frequency details essential for small object detection. To mitigate this, we propose the Content Reconstruction Upsampler. Leveraging content-aware feature reassembly~\cite{wang2019carafe}, this module predicts dynamic kernels guided by local semantics to effectively restore object details during upsampling. The process is defined as:
\begin{equation}
F_{\text{out}}(i, j) = \sum_{(p,q) \in \mathcal{N}_k(i,j)} K_{i,j}(p,q) \cdot F_{\text{in}}(p,q)
\label{eq:carafe}
\end{equation}
where $\mathcal{N}_k(i, j)$ is the $k \times k$ neighborhood, and $K_{i,j}$ is the predicted reassembly kernel.

\noindent\textbf{2) Geometric Alignment Downsampler.} Conversely, shallow features often contain geometric distortions caused by varying UAV viewpoints, leading to spatial misalignment. To address this, we introduce the Geometric Alignment Downsampler. By incorporating adaptive deformable sampling~\cite{zhu2019deformable}, this module augments the sampling grid with learned 2D offsets and modulation factors. This enables the network to model geometric variations and align features precisely during downsampling. The process is formulated as:
\begin{equation}
F_{\text{out}}(p_0) = \sum_{p_n \in \mathcal{R}} w(p_n) \cdot F_{\text{in}}(p_0 + p_n + \Delta p_n) \cdot m_n
\label{eq:dcnv2}
\end{equation}
where $\mathcal{R}$ is the regular grid, while $\Delta p_n$ and $m_n$ are the learned sampling offsets and modulation scalars.

\noindent\textbf{3) Adaptive Weighted Fuser.} To effectively integrate the refined features with mid-level representations, we introduce the Adaptive Weighted Fuser. Utilizing a fast normalized weighted sum strategy~\cite{tan2020efficientdet}, this module dynamically learns contribution weights for the three inputs ($F'_{\text{deep}}$, $F_{\text{mid}}$, and $F'_{\text{shallow}}$). This mechanism enables the network to adaptively emphasize salient information while suppressing noise. The fusion is defined as:
\begin{equation}
F_{\text{fused}} = \frac{w_1 F'_{\text{deep}} + w_2 F_{\text{mid}} + w_3 F'_{\text{shallow}}}{w_1 + w_2 + w_3 + \epsilon}
\label{eq:bifpn_fusion}
\end{equation}
where $w_i$ are learnable scalar weights and $\epsilon$ is a small constant for numerical stability.

\noindent\textbf{4) Architecture and Prediction.} By stacking ASFBs within bidirectional top-down and bottom-up paths, HFAN generates a feature pyramid $\{P'_2, P'_3, P'_4, P'_5\}$ rich in both semantic and spatial information. To fully exploit the high-resolution $P'_2$ feature map, we introduce a dedicated fourth detection head tailored for small object detection. Consistent with modern practices~\cite{Github'24}, all detection heads employ a decoupled architecture, separating classification and bounding box regression tasks.

\section{Experiments}

\subsection{Experimental Setup}

\textbf{Datasets.} We conduct experiments on two distinct, challenging UAV datasets to comprehensively validate our approach. 
\begin{itemize}
    \item \textbf{DroneVehicle:} A large-scale dual-modality (RGB-IR) dataset for vehicle detection. We use it to evaluate the performance of our complete MambaRefine-YOLO model.
    \item \textbf{VisDrone:} A large-scale, highly challenging single-modality (RGB) dataset. We use it to test the generalization capability and independent effectiveness of our proposed HFAN.
\end{itemize}

\textbf{Metrics and Details.} We use mean Average Precision (mAP@.5) as the primary accuracy metric. We also report parameters (Params) for efficiency analysis. All models were implemented in PyTorch 2.0.0 and trained on single NVIDIA RTX 4090 GPU using the AdamW optimizer for 300 epochs with an input size of $640 \times 640$.

\subsection{Evaluation on Dual-Modality Data (DroneVehicle)}

In this section, we evaluate the full MambaRefine-YOLO model on the DroneVehicle dataset to demonstrate the effectiveness of our dual-modality strategy.

\textbf{Comparison with State-of-the-Art.} As shown in Table \ref{tab:dronevehicle_sota_comprehensive}, our model is compared with various dual-modality detectors. While single-modality YOLO11 baselines show limited performance, our MambaRefine-YOLO achieves a superior 83.5\% mAP. Notably, it outperforms the Transformer-based CFT~\cite{fang2021_cft} by a significant \textbf{6.3\%} margin and surpasses the recent SOTA UAVD-Mamba~\cite{Li2025UAVD-Mamba}, demonstrating the robust effectiveness of our Mamba-based fusion strategy.

\begin{table}[!t]
\centering
\caption{Ablation study of MambaRefine-YOLO's key components on the DroneVehicle dataset. 
         ``Concat'' denotes simple feature concatenation, ``Uni-Mamba'' and ``Bi-Mamba'' correspond to unidirectional and bidirectional Mamba-based fusion, respectively, and ``+ DGC-Gate'' indicates enabling our Dual-Gated Complementary Mamba fusion module (DGC-MFM). Best results are in bold.}
\label{tab:comprehensive_ablation}
\resizebox{\columnwidth}{!}{%
\begin{tabular}{c|cccc|cc}
\toprule
\multirow{2}{*}{Neck} & \multicolumn{4}{c|}{Backbone Fusion Strategy} & \multirow{2}{*}{mAP@.5} & \multirow{2}{*}{$\Delta$} \\
\cmidrule{2-5} 
 & Concat & Uni-Mamba & + Bi-Mamba & + DGC-Gate & & \\
\midrule
\midrule
FPN & \checkmark & & & & 79.4 & Ref. \\
\textbf{HFAN} & \checkmark & & & & 80.5 & +1.1 \\
\midrule
FPN & & \checkmark & & & 81.0 & +1.6 \\
FPN & & & \checkmark & & 81.8 & +0.8 \\
FPN & & & \checkmark & \checkmark & 82.3 & +0.5 \\
\midrule
\textbf{HFAN} & & & \checkmark & \checkmark & \textbf{83.2} & \textbf{+3.8} \\
\bottomrule
\end{tabular}%
}
\end{table}

\begin{table}[!t]
\centering
\small 
\renewcommand{\arraystretch}{0.9} 
\caption{Comparison with state-of-the-art detectors on the VisDrone dataset. 
         Our model achieves a highly competitive performance-to-parameter trade-off. 
         Best results are in \textbf{bold}.}
\label{tab:visdrone_sota_compact}
\begin{tabular}{l c c}
\toprule
\renewcommand{\arraystretch}{0.9} 
\textbf{Model} & \textbf{AP$_{50}$ (\%)} & \textbf{Params (M)} \\
\midrule
\midrule
YOLOv8m \cite{Github'24} & 40.3 & 25.09 \\
YOLOv12m \cite{Github'24}& 41.2 & 19.67 \\
\midrule
DAB-Deformable-DETR \cite{hic_yolov5_ref} & 44.2 & 47.06 \\
DEIM \cite{deim_ref} & 45.6 & 19.19 \\
QueryDet \cite{querydet_ref} & 48.1 & 33.9 \\
DCFL \cite{clusdet_ref} & 36.1 & 32.1 \\
\midrule
\textbf{HFAN-YOLO-S} & 44.8 & 13.09 \\
\textbf{HFAN-YOLO-M} & \textbf{49.4} & 25.66 \\ 
\bottomrule
\end{tabular}
\end{table}

\textbf{Ablation Study on Components.} To dissect the contributions of our two main innovations, we conducted an ablation study on DroneVehicle, presented in Table \ref{tab:comprehensive_ablation}. The baseline model uses simple concatenation for fusion and a standard FPN. 
Replacing the standard FPN with our HFAN yields a mAP gain of \textbf{+1.1\%} (80.5\%), proving the efficacy of our hierarchical aggregation. 
On the fusion side, replacing concatenation with Uni-Mamba improves performance to 81.0\%, and upgrading to the full DGC-Gate Mamba further boosts it. 
Finally, when both the full fusion module and HFAN are used together, the model achieves the best result of \textbf{83.2\%}, demonstrating their synergistic effect.

\subsection{Generalization on Single-Modality Data (VisDrone)}

To verify the independent value and general applicability of our proposed HFAN, we adapted our model for a single-modality task. We constructed a variant, named HFAN-YOLO, by replacing the standard FPN neck in YOLOv8 with our HFAN, while retaining the original single-stream backbone. We evaluated Small (S) and Medium (M) scales on the challenging VisDrone dataset.

\textbf{Comparison with State-of-the-Art.} Table \ref{tab:visdrone_sota_compact} compares HFAN-YOLO with other SOTA detectors. Our HFAN-YOLO-M achieves \textbf{49.4\%} mAP, significantly outperforming the baseline YOLOv8m (40.3\%) and the recent YOLOv12m (41.2\%). Moreover, it surpasses specialized end-to-end detectors like \cite{deim_ref} (45.6\%) and QueryDet \cite{querydet_ref} (48.1\%) while maintaining a favorable parameter count (25.66M). This demonstrates that HFAN serves as a powerful, plug-and-play module that substantially enhances detection performance without excessive computational cost.

\textbf{Ablation Study on HFAN.} We conducted a detailed ablation study on VisDrone to analyze the internal components of HFAN, as shown in Table \ref{tab:hfan_ablation_final}. Note that for efficiency, this analysis was performed on the HFAN-YOLO-N scale. Comparing the Full Model with the FPN Baseline, our HFAN delivers a substantial \textbf{+6.0\%} improvement. To validate specific modules, we employ a ``remove-one'' strategy: removing the Hierarchical Structure (w/o HS) causes the largest drop of \textbf{3.6\%}, underscoring the importance of our multi-scale aggregation. Similarly, excluding the Content Reconstruction Upsampler (w/o CRU), the Geometric Alignment Downsampler (w/o GAD) or the Adaptive Weighted Fuser (w/o AWF) leads to performance declines of \textbf{1.9\%}, \textbf{2.0\%} and \textbf{1.1\%} respectively, confirming that all components are essential to the ``Refine-then-Fuse'' design.

\begin{table}[!t]
\centering
\caption{Ablation study of the core components
         on the VisDrone dataset. The drop ($\downarrow$) is calculated relative to the Full Model.}
\label{tab:hfan_ablation_final}
\resizebox{\columnwidth}{!}{%
\begin{tabular}{l|cccc|c|c}
\toprule
\multirow{2}{*}{Configuration} & \multicolumn{4}{c|}{Components} & \multirow{2}{*}{mAP@.5} & \multirow{2}{*}{Drop ($\downarrow$)} \\
\cmidrule{2-5} 
 & HS & CRU & GAD & AWF & & \\
\midrule
\midrule
\textbf{HFAN-YOLO} & \checkmark & \checkmark & \checkmark & \checkmark & \textbf{39.1} & - \\
\midrule
w/o HS (Structure only) & & \checkmark & \checkmark & \checkmark & 35.5 & 3.6 \\
w/o CRU & \checkmark & & \checkmark & \checkmark & 37.2 & 1.9 \\
w/o GAD & \checkmark & \checkmark & & \checkmark & 37.1 & 2.0 \\
w/o AWF & \checkmark & \checkmark & \checkmark & & 38.0 & 1.1 \\
\midrule
FPN (Baseline) & & & & & 33.1 & 6.0 \\
\bottomrule
\end{tabular}%
}
\end{table}

\begin{figure}[!t]
\centering
\includegraphics[width=\columnwidth]{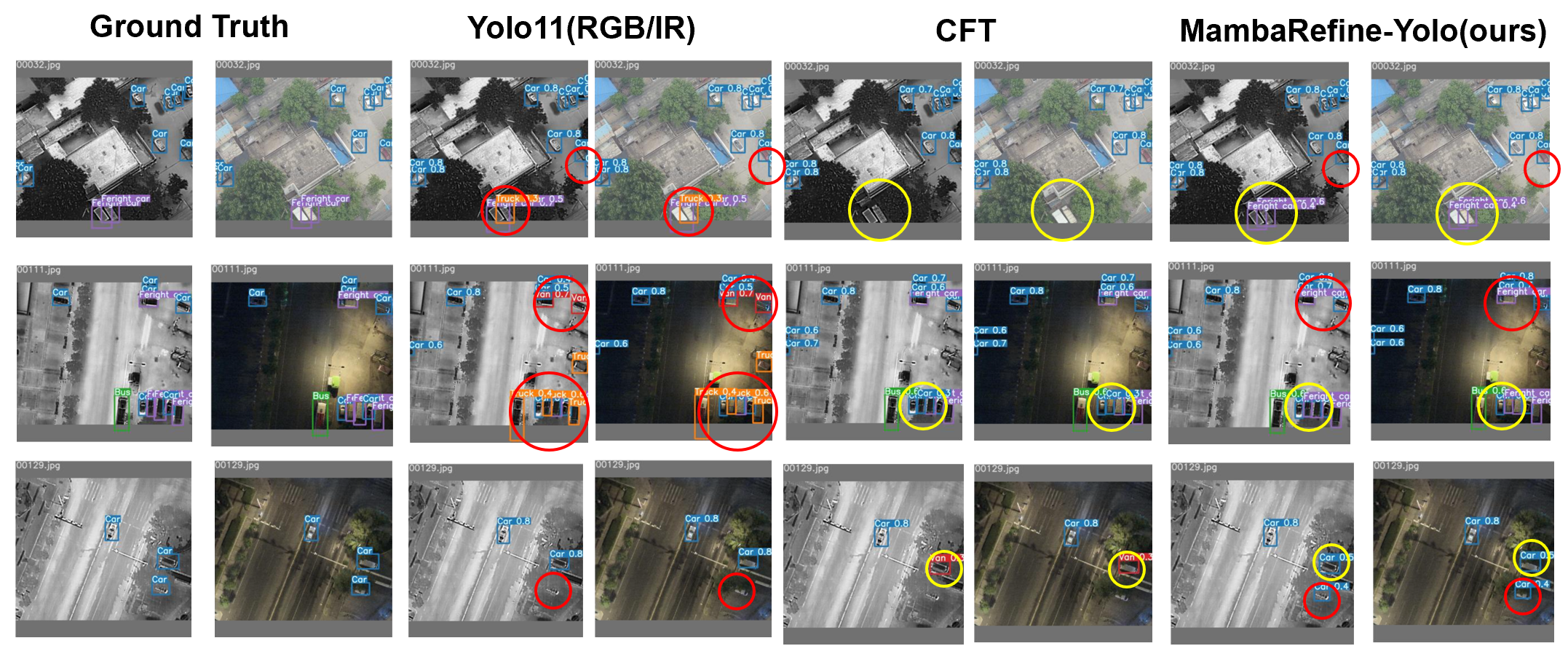}
\caption{Qualitative Results of MambaRefine-YOLO (ours) vs. SOTA Methods on DroneVehicle (RGB/IR Fusion). Red and yellow circles highlight misdetections.}
\label{fig:results}
\end{figure}

\subsection{Qualitative Results}
Fig. \ref{fig:results} visualizes the detection performance on the DroneVehicle dataset compared to Ground Truth, YOLO11, and CFT. As highlighted by the red circles, the baseline YOLO11 suffers from missed detections and false positives, particularly in low-light regions. While the Transformer-based CFT improves stability, it still fails to recall specific targets like freight cars and trucks. In contrast, our MambaRefine-YOLO effectively corrects these errors. By leveraging robust cross-modal features, it accurately detects small and obscured objects where other methods fail, aligning closely with the Ground Truth.

\vspace{-0.2cm}
\section{Conclusion}
In this letter, we proposed MambaRefine-YOLO, a novel framework that synergizes the Dual-Gated Complementary Mamba Fusion Module (DGC-MFM) and the Hierarchical Feature Aggregation Neck (HFAN). Extensive experiments validate the distinct value of these contributions: the DGC-MFM establishes a new state-of-the-art on the dual-modality DroneVehicle dataset (83.2\% mAP), significantly outperforming Transformer-based alternatives in complex illumination conditions. Meanwhile, the standalone HFAN demonstrates remarkable generalization on the single-modality VisDrone dataset, surpassing recent detectors like YOLOv12. By effectively balancing cross-modal interaction and small object refinement, our approach offers a superior trade-off between accuracy and efficiency, making it an ideal solution for real-time UAV surveillance.

\vspace{-0.5cm}

\bibliographystyle{IEEEtran}
\bibliography{references}

\end{document}